\documentclass[10pt,twocolumn,letterpaper]{article}

\usepackage[pagenumbers]{cvpr} 

\definecolor{cvprblue}{rgb}{0.21,0.49,0.74}
\usepackage[pagebackref,breaklinks,colorlinks,allcolors=cvprblue]{hyperref}

\newcommand{\tabincell}[2]{\begin{tabular}{@{}#1@{}}#2\end{tabular}}
\usepackage{multirow}
\usepackage{threeparttable}
\usepackage{bbm}
\usepackage{float}
\usepackage{booktabs} 
\usepackage{adjustbox}

\def\paperID{2173} 
\def\confName{CVPR}
\def\confYear{2025}

\title{How Texts Help? A Fine-grained Evaluation to Reveal the Role of Language in Vision-Language Tracking}

\author{\textbf{Xuchen Li}$^{1,2}$\thanks{Equal contribution.} \hspace{9pt} 
\textbf{Shiyu Hu}$^{3*}$\hspace{9pt}
\textbf{Xiaokun Feng}$^{1,2}$ \hspace{9pt}
\textbf{Dailing Zhang}$^{1,2}$\hspace{9pt}
\\
\textbf{Meiqi Wu}$^4$ \hspace{9pt}
\textbf{Jing Zhang}$^1$ \hspace{9pt}
\textbf{Kaiqi Huang}$^{1,2,5}$ \hspace{9pt}\\
\textsuperscript{1}CRISE, Institute of Automation, Chinese Academy of Sciences\\
\textsuperscript{2}School of Artificial Intelligence, University of Chinese Academy of Sciences\\
\textsuperscript{3}School of Physical and Mathematical Sciences, Nanyang Technological University\\
\textsuperscript{4}School of Computer Science and Technology, University of Chinese Academy of Sciences\\
\textsuperscript{5}CAS Center for Excellence in Brain Science and Intelligence Technology\\
\tt\small lixuchen2024@ia.ac.cn, shiyu.hu@ntu.edu.sg, fengxiaokun2022@ia.ac.cn\\
\tt\small zhangdailing2023@ia.ac.cn, wumeiqi18@mails.ucas.ac.cn\\
\tt\small jing\_zhang@ia.ac.cn, kaiqi.huang@nlpr.ia.ac.cn
}

\begin{document}
\maketitle
\begin{abstract}

Vision-language tracking (VLT) extends traditional single object tracking by incorporating textual information, providing semantic guidance to enhance tracking performance under challenging conditions like fast motion and deformations. However, current VLT trackers often underperform compared to single-modality methods on multiple benchmarks, with semantic information sometimes becoming a “distraction.” To address this, we propose \textbf{VLTVerse}, the first fine-grained evaluation framework for VLT trackers that comprehensively considers multiple challenge factors and diverse semantic information, hoping to reveal the role of language in VLT. Our contributions include: (1) VLTVerse introduces 10 sequence-level challenge labels and 6 types of multi-granularity semantic information, creating a flexible and multi-dimensional evaluation space for VLT; (2) leveraging 60 subspaces formed by combinations of challenge factors and semantic types, we conduct systematic fine-grained evaluations of three mainstream SOTA VLT trackers, uncovering their performance bottlenecks across complex scenarios and offering a novel perspective on VLT evaluation; (3) through decoupled analysis of experimental results, we examine the impact of various semantic types on specific challenge factors in relation to different algorithms, providing essential guidance for enhancing VLT across data, evaluation, and algorithmic dimensions. The VLTVerse, toolkit, and results will be available at \url{http://metaverse.aitestunion.com}.

\end{abstract}    
\section{Introduction}
\label{sec:introduction}
\begin{figure}[htbp!]
  \centering
   \vspace{-18pt}
   \includegraphics[width=1\linewidth]{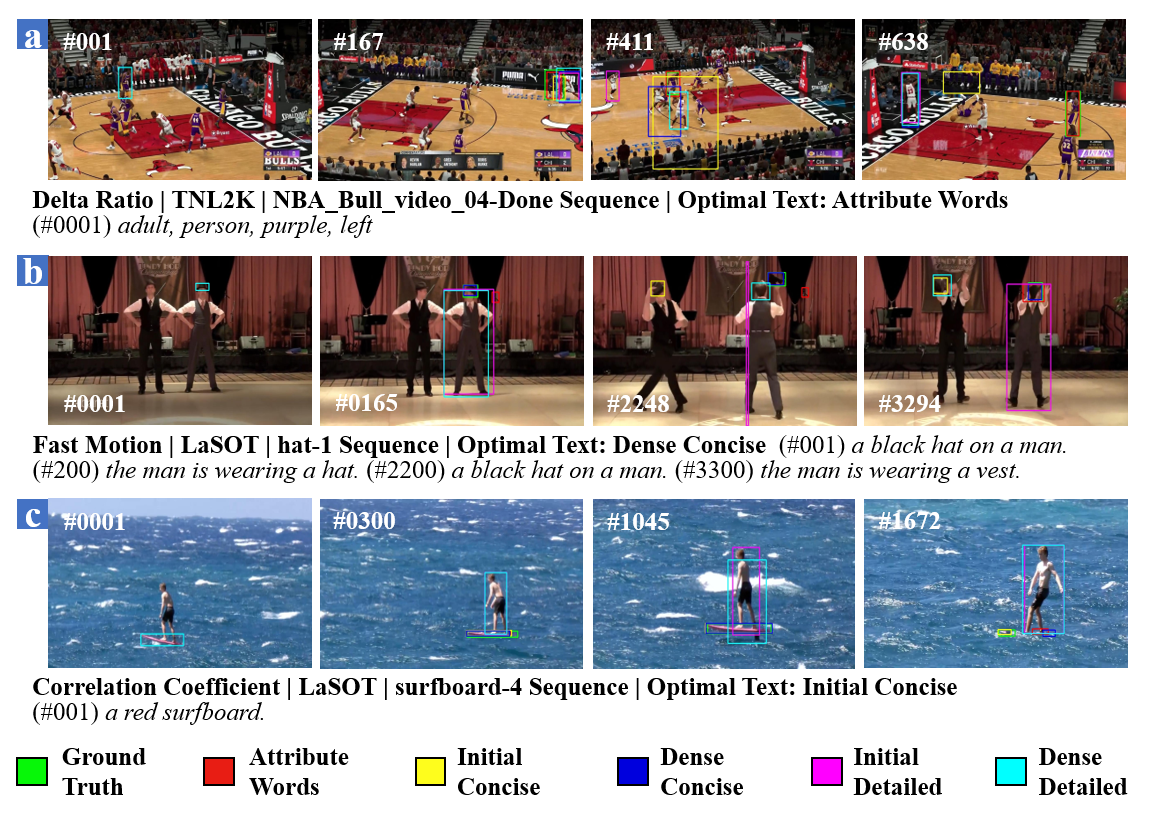}
   \vspace{-8pt}
   \caption{Examples of tracking results by JointNLT \cite{jointnlt} under sequences with three different challenge factors combined with various texts. We select representative sequences under the three most challenging factors (delta ratio, fast motion, and correlation coefficient) from two tracking datasets (LaSOT \cite{lasot} and TNL2K \cite{tnl2k}). It is evident that the tracking performance of JointNLT varies significantly with different textual assistance, and the figure labels the text that results in the best tracking performance. Faced with different challenge factors, different semantic information might be needed to provide guidance. Otherwise, the text could become a distraction. VLTVerse reveals the shortcomings of traditional evaluation methods and offers guidance for tracker optimization from the perspective of fine-grained evaluation.}
   \label{fig:motivation}
   \vspace{-12pt}
\end{figure}

Vision-Language Tracking (VLT) advances traditional Single Object Tracking (SOT) by integrating high-level semantic information from language \cite{otb99}, complementing the visual modality and aiming to enhance tracking performance, particularly under challenging conditions.

In practical scenarios, tracking tasks often encounter various challenge factors, such as fast motion, deformations, and lighting variations, which disrupt the appearance and motion information in the visual modality, significantly impairing tracking performance \cite{sotverse}. This degradation of visual information makes it difficult for models to rely solely on visual cues for stable tracking under challenging conditions. In this context, language information offers a potential supplement for VLT tasks. When visual information is compromised, precise and appropriate language descriptions can provide semantic support \cite{dtvlt,dtllm-vlt}, helping the model maintain robustness in the face of challenges.

However, the diversity and flexibility of information in the language can also introduce potential distractions. If the language information is imprecise or inconsistent with the visual data, it may further degrade the tracker’s performance \cite{dtvlt}. This dual role of language information makes its effective utilization a central issue in VLT research. In fact, most existing VLT benchmarks provide only one style of textual annotations for the language modality and lack fine-grained annotations and analyses of challenge factors for the visual modality, as seen in visual tracking benchmarks. Consequently, they fail to adequately consider the diversity of challenge factors and the dual impact of language information, limiting their ability to assess VLT tracker in complex scenarios. As shown in Figure \ref{fig:motivation}, under the typical sequences of the three challenging factors, variations in the text significantly affect the tracking performance of the JointNLT \cite{jointnlt}, and the text that achieves optimal tracking performance is not the same. This raises several key questions: What types of textual information can effectively supplement visual cues under challenging conditions? How does the introduction of language impact tracker performance in different contexts? To address these questions, a framework capable of fine-grained evaluation to reveal the role of language is required.

To this end, we propose VLTVerse, the first fine-grained evaluation framework for VLT, designed to overcome the limitations of existing datasets and enhance VLT system performance and evaluation capabilities. Building on SOTVerse \cite{sotverse}, VLTVerse further expands the evaluation dimensions to encompass 10 sequence-level challenge factors and 6 types of multi-granularity semantic information. By combining these factors and language information, VLTVerse creates an evaluation space of 60 combinations of challenges and semantics, enabling systematic fine-grained evaluation of mainstream VLT trackers across diverse scenarios. Researchers can select appropriate data (including challenge factors and textual information) and evaluation metrics to identify performance bottlenecks and optimize algorithm designs. By providing a fine-grained evaluation space for VLT tasks, VLTVerse not only reveals the potential supplementary role of language across various challenge factors but also addresses the limitations of traditional evaluation methods, offering valuable guidance for the future development of VLT research.

The contributions of this paper can be summarized as follows:

\begin{itemize}
\item We construct VLTVerse, the first fine-grained evaluation framework for VLT, covering short-term, long-term, and global instance tracking tasks. This framework includes four representative benchmarks, ten challenge factors, and six types of semantic information, enabling comprehensive evaluation of VLT trackers. (Sec. \ref{sec:vltverse} and Sec. \ref{sec:environment})

\item For the first time, we combine ten challenge factors with six types of semantic information in the VLT task and conduct systematic performance evaluations of mainstream VLT trackers across 60 combinations. This fine-grained evaluation reveals critical performance insights on the role of language that traditional evaluation methods cannot capture. (Sec. \ref{sec:evaluation} and Sec. \ref{sec:executor})

\item We provide an in-depth analysis of the language modality’s impact on VLT trackers, particularly under various challenge factors. Through VLTVerse’s fine-grained evaluation, we enhance our understanding of VLT tasks and offer valuable guidance for improving tracker performance in future research. (Sec. \ref{sec:analysis})

\end{itemize}
\section{Related Work}
\label{sec:relatedwork}

\subsection{Vision-Language Tracking Benchmark}
As VLT research progresses, multiple benchmarks have been introduced to advance the field. Early VLT benchmarks primarily added semantic annotations to existing single object tracking benchmarks. OTB99\_Lang \cite{otb99}, the first VLT benchmark, extended the OTB \cite{otb100,otb50} dataset with natural language descriptions, pioneering this new task. To address the need for larger datasets in VLT research, LaSOT \cite{lasot,lasotext} expanded the task to long-term tracking, contributing substantially to VLT's evolution. In the same year, TNL2K \cite{tnl2k} was introduced specifically for VLT, aiming to improve tracking flexibility and accuracy through detailed textual descriptions. Subsequently, the MGIT \cite{mgit} benchmark broadened VLT’s scope to global instance tracking \cite{git}, supporting a deeper understanding of video content with rich spatiotemporal and causal relationships. Recently, the large-scale dataset Elysium-1M \cite{elysium-1m} was released, supporting three distinct VLT tasks. Additionally, several benchmarks have begun focusing on specific scenarios, such as tracking in wild \cite{semtrack}, underwater \cite{webuot-1m,uw-cot}, and drone environments \cite{webuav-3m}. These benchmarks collectively enrich the dataset landscape and drive the development of diverse VLT trackers. However, most VLT benchmarks still rely on semantic extensions of SOT datasets and lack unified annotation standards, often providing insufficiently detailed descriptions for challenging tracking scenarios.

\subsection{Vision-Language Tracking Algorithm}
The VLT algorithm extends SOT to the multi-modal domain by integrating language descriptions with an initial template frame. Most VLT trackers \cite{vlt,transnlt,transvlt,ctrtnl,arxiv19,dat,rttnld,osdt,ovlm,snlt} achieve tracking through similarity matching, aligning language descriptions and template frames to identify the best-matching target in the search frames. MMTrack \cite{mmtrack} enhances multi-modal understanding by reframing the VLT task as a token generation problem using unified token learning. Additionally, several trackers leverage temporal information to boost tracking performance. For instance, GTI \cite{gti} and AdaSwitcher \cite{tnl2k} integrate tracking and localization outputs to improve accuracy, while MemVLT \cite{memvlt} achieves robust tracking via a short- and long-term memory interaction mechanism with adaptive prompts. QueryNLT \cite{querynlt} maintains temporal consistency by utilizing historical visual information, enabling precise tracking across frames. Recent models \cite{elysium-1m,attracker,chattracker} explore semi-supervised methods or large language model integration to enhance VLT. Some trackers also aim to support multiple tasks within a unified model: JointNLT \cite{jointnlt} combines temporal information to improve both visual grounding and VLT performance, while UVLTrack \cite{uvltrack} employs a unified feature extractor to handle VLT, visual grounding, and SOT simultaneously. Despite these advancements, VLT trackers still fall short of state-of-the-art (SOTA) SOT trackers \cite{rtracker,aqatrack,odtrack,evptrack,hiptrack}, underscoring the need for comprehensive and fine-grained evaluation of current VLT approaches.

\subsection{Vision-Language Tracking Evaluation}
Significant advancements have been made in the evaluation methods and techniques within the SOT field, beginning with the introduction of novel evaluation metrics through benchmarks like OTB \cite{otb50,otb100} and competitions such as VOT \cite{vot18,vot19,vot20,vot21,vot22,vot-tir15,vot16}, which have gradually become standard in SOT tasks. The emergence of SOTVerse \cite{sotverse} has further accelerated this development by integrating environment, evaluation, and executor components, allowing users to create custom SOT spaces and perform in-depth analyses on various challenge factors. This approach uncovers limitations in existing trackers and offers a more flexible and comprehensive means of dataset utilization for evaluation. In contrast, evaluation methods for VLT tasks remain relatively underdeveloped. Although some initial efforts have been made—such as VLT-MI \cite{vlt-mi}, which proposes metrics for assessing multi-round interactions, and DTVLT \cite{dtvlt,dtllm-vlt}, which evaluates VLT trackers through multi-granularity textual metrics—these methods lack the comprehensiveness and detail needed to address the specific challenge factors encountered during tracking. Tackling these challenge factors is essential to fundamentally enhance VLT performance, necessitating a more exhaustive evaluation framework that can help researchers identify and overcome current limitations, thereby providing valuable insights for future research.

\begin{figure}[t!]
  \centering
  \vspace{-20pt}
    \includegraphics[width=1\linewidth]{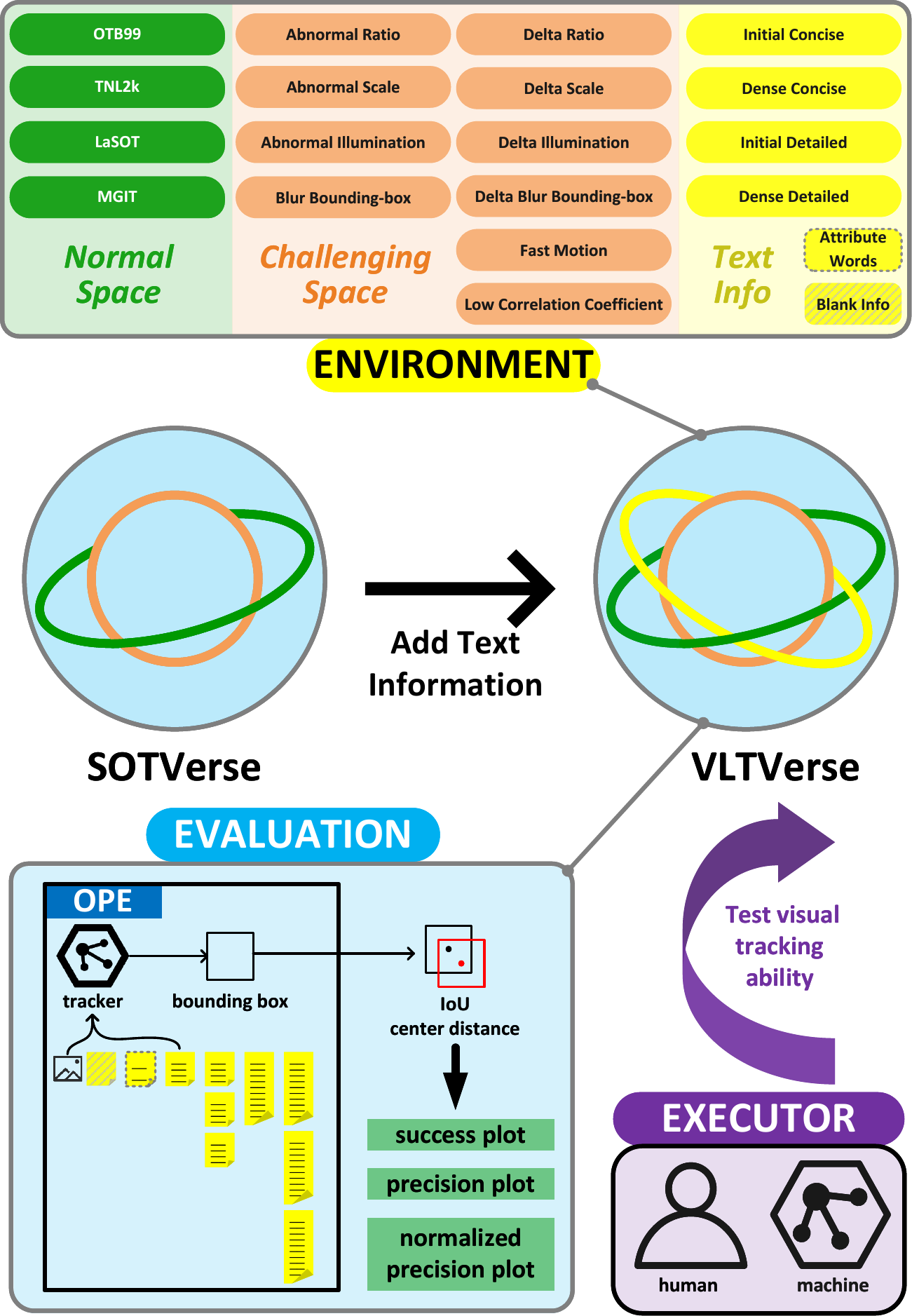}
    \vspace{-8pt}
    \caption{VLTVerse comprises two main components: environment and evaluation. As an extension of SOTVerse \cite{sotverse}, it expands the evaluation space into three dimensions—normal space, challenge factor space, and textual information space. The normal space covers short-term, long-term, and global instance tracking tasks. The challenge factor space is defined by 10 attributes corresponding to 10 distinct challenge factors, while the textual information space includes 6 types of semantic descriptions. This three-dimensional framework enables a comprehensive evaluation of tracking performance under various language and challenge conditions. Using the OPE evaluation system, we assess tracker performance across different challenge factor spaces with diverse textual inputs. Key evaluation metrics include SR, SUC, PRE, N-PRE, and AUC. Based on the defined environment and evaluation setup, researchers can design customized executors by combining specific textual information and challenge factors, thus creating experimental settings that allow for a fine-grained analysis of language's role in VLT.}
    \label{fig:3e_paradigm}
    \vspace{-12pt}
\end{figure}
\section{VLTVerse}
\label{sec:vltverse}

SOTVerse \cite{sotverse} introduces the 3E paradigm, dividing tasks into three components: Environment, Evaluation, and Executor. VLTVerse also adopts the 3E paradigm but further expands the Environment dimension to encompass \textit{Normal Space}, \textit{Challenge Factor Space}, and \textit{Textual Information Space}. Let $S$ denote a subtask (e.g., short-term tracking task), $E$ the corresponding experimental environment (e.g., short-term dataset), $M_s$ the evaluation system (e.g., OPE mechanism), $M_m$ the evaluation metrics (e.g., precision), $T$ the set of task executors (e.g., VLT trackers), and $\times$ the Cartesian product. Under the 3E paradigm, the subtask is represented as:

\begin{align}
S = E \times M_s \times M_m \times T
\end{align}

A complete VLT task, $\mathbb{S}$, consists of multiple subtasks. In VLTVerse, we decompose $\mathbb{S}$ into 60 subtasks $S_{ci}$ ($S_{c} \times S_{i}$), where $S_c$ ($S_{c1}$ - $S_{c10}$) represents challenge factor subtasks, and $S_i$ ($S_{i1}$ - $S_{i6}$) denotes textual information subtasks. The corresponding environments are defined as $E_c$, $E_i$, and $E_{ci}$ for each respective subtask in $S$. We denote the environment, evaluation, and executor components as $\mathbb{E}$, $\mathbb{M}$, and $\mathbb{T}$, respectively. Thus, the task $\mathbb{S}$ and its environment $\mathbb{E}$ are defined as follows

\begin{align}
\mathbb{S} &= \! \{S_{c1i1} , \ldots , S_{c1i6} , \ldots , S_{c10i1} , \ldots\} \! = \mathbb{E} \times \mathbb{M} \times \mathbb{T} \\
\mathbb{E} &= \! \{E_{c1i1} , \ldots , E_{c1i6} , \ldots , E_{c10i1} , \ldots\} \! = E_c \times E_i
\end{align}

Based on the 3E Paradigm, we develop VLTVerse, a fine-grained evaluation framework that addresses diverse textual and scenario challenges by integrating existing VLT datasets into an expansive environmental space, $\mathbb{E}$. VLTVerse provides an OPE mechanism with multiple metrics, $\mathbb{M}$, enabling a comprehensive assessment of language function for VLT trackers, $\mathbb{T}$. Utilizing VLTVerse, researchers can efficiently extract relevant subtasks, configure the environment, and select VLT trackers for performance evaluation.

Similar to SOTVerse \cite{sotverse}, VLTVerse also allows researchers to define the task space through interaction and extension. Researchers can interactively retrieve data, set evaluation metrics, and analyze experimental outcomes. Through extensions, they can add new data or metrics to conduct detailed evaluations of tracker performance across varying tasks, focusing on the role of language. For further details on VLTVerse, please refer to Appendix A.
\section{Environment}
\label{sec:environment}

\begin{table*}[ht]
\centering
\vspace{-20pt}
\small
\renewcommand{\arraystretch}{1.2}
\setlength{\tabcolsep}{5pt}
\begin{adjustbox}{width=1\textwidth}
\begin{tabular}{@{}l|cccccccc@{}}
\toprule
\multirow{3}{*}{\textbf{Challenge Space}} & \multicolumn{2}{c}{\textbf{OTB99\_Lang \cite{otb99}}} & \multicolumn{2}{c}{\textbf{TNL2K \cite{tnl2k}}} & \multicolumn{2}{c}{\textbf{LaSOT \cite{lasot}}} & \multicolumn{2}{c}{\textbf{MGIT \cite{mgit}}} \\
\cmidrule(lr){2-3} \cmidrule(lr){4-5} \cmidrule(lr){6-7} \cmidrule(lr){8-9}
& \textbf{\# Video} & \textbf{Mean Frame} & \textbf{\# Video} & \textbf{Mean Frame} & \textbf{\# Video} & \textbf{Mean Frame} & \textbf{\# Video} & \textbf{Mean Frame} \\
& \textbf{(Train / Test)} & \textbf{(Train / Test)} & \textbf{(Train / Test)} & \textbf{(Train / Test)} & \textbf{(Train / Test)} & \textbf{(Train / Test)} & \textbf{(Train / Test)} & \textbf{(Train / Test)} \\
\midrule
\textbf{$E_{c1}$ (Abnormal Ratio)} & 37 / 33 & 621 / 583 & 1,058 / 622 & 532 / 738 & 1,086 / 271 & 2,526 / 2,457 & 84 / 12 & 12,333 / 13,051 \\
\textbf{$E_{c2}$ (Abnormal Scale)} & 49 / 47 & 573 / 616 & 1,156 / 619 & 551 / 734 & 898 / 254 & 2,555 / 2,461 & 88 / 12 & 13,047 / 13,088 \\
\textbf{$E_{c3}$ (Blur Bounding-box)} & 1 / 0 & 71 / 0 & 235 / 150 & 704 / 847 & 214 / 53 & 2,286 / 2,305 & 60 / 9 & 12,290 / 10,819 \\
\textbf{$E_{c4}$ (Abnormal Illumination)} & 23 / 24 & 523 / 653 & 666 / 458 & 583 / 775 & 864 / 220 & 2,518 / 2,482 & 79 / 11 & 14,137 / 12,548 \\
\hline
\textbf{$E_{c5}$ (Delta Illumination)} & 13 / 12 & 305 / 311 & 404 / 290 & 544 / 800 & 178 / 44 & 2,419 / 2,122 & 34 / 4 & 10,322 / 11,671 \\
\textbf{$E_{c6}$ (Delta Scale)} & 1 / 8 & 500 / 406 & 206 / 121 & 478 / 559 & 19 / 5 & 2,369 / 2,734 & 14 / 2 & 7,690 / 5,331 \\
\textbf{$E_{c7}$ (Delta Blur)} & 20 / 18 & 333 / 610 & 121 / 35 & 474 / 544 & 73 / 27 & 2,714 / 2,440 & 2 / 0 & 6,919 / 0 \\
\textbf{$E_{c8}$ (Delta Ratio)} & 5 / 4 & 391 / 511 & 75 / 51 & 330 / 430 & 5 / 0 & 2,280 / 0 & 2 / 0 & 14,136 / 0 \\
\textbf{$E_{c9}$ (Fast Motion)} & 9 / 10 & 184 / 393 & 45 / 34 & 525 / 707 & 102 / 23 & 2,854 / 2,398 & 17 / 1 & 15,226 / 21,421 \\
\textbf{$E_{c10}$ (Correlation Coefficient)} & 5 / 5 & 110 / 312 & 32 / 13 & 411 / 337 & 32 / 7 & 2,185 / 1,751 & 3 / 0 & 15,114 / 0 \\
\bottomrule
\end{tabular}
\end{adjustbox}
\vspace{-8pt}
\caption{Statistics of challenge factors across datasets.}
\label{tab:challengingfactor}
\vspace{-12pt}
\end{table*}

\begin{table}[ht]
\centering
\small
\renewcommand{\arraystretch}{1.2}
\setlength{\tabcolsep}{6pt}
\begin{adjustbox}{width=1\linewidth}
\begin{tabular}{@{}c|ccccccc@{}}
\toprule
\textbf{\tabincell{c}{Challenge\\Space}} & \textbf{\tabincell{c}{Attribute\\Words \\ $E_{i1}$}} & \textbf{\tabincell{c}{Dense\\Concise \\ $E_{i2}$}} & \textbf{\tabincell{c}{Dense\\Detailed \\ $E_{i3}$}} & \textbf{\tabincell{c}{Initial\\Concise \\ $E_{i4}$}} & \textbf{\tabincell{c}{Initial\\Detailed \\ $E_{i5}$}} & \textbf{\tabincell{c}{Blank \\ $E_{i6}$}} \\
\midrule
\textbf{$E_{c1}$} & 4 & 76 & 485 & 7 & 29 & 3 \\
\textbf{$E_{c2}$} & 4 & 73 & 463 & 7 & 28 & 3 \\
\textbf{$E_{c3}$} & 4 & 106 & 674 & 8 & 29 & 3 \\
\textbf{$E_{c4}$} & 4 & 87 & 558 & 8 & 29 & 3 \\
\hline
\textbf{$E_{c5}$} & 4 & 58 & 373 & 7 & 28 & 3 \\
\textbf{$E_{c6}$} & 4 & 40 & 261 & 6 & 27 & 3 \\
\textbf{$E_{c7}$} & 4 & 53 & 327 & 7 & 27 & 3 \\
\textbf{$E_{c8}$} & 4 & 27 & 175 & 5 & 24 & 3 \\
\textbf{$E_{c9}$} & 4 & 127 & 786 & 8 & 29 & 3 \\
\textbf{$E_{c10}$} & 4 & 72 & 477 & 5 & 28 & 3 \\
\bottomrule
\end{tabular}
\end{adjustbox}
\vspace{-8pt}
\caption{Statistics of textual information across factors based on mean word count.}
\label{tab:information}
\vspace{-12pt}
\end{table}

\subsection{Dataset Selection}

We select representative datasets from short-term, long-term, and global instance tracking tasks to ensure VLTVerse effectively reflects the diversity and characteristics of VLT tasks. The chosen datasets—OTB99\_Lang \cite{otb99}, TNL2K \cite{tnl2k}, LaSOT \cite{lasot}, and MGIT \cite{mgit}—collectively span 6.85 million frames, enabling a comprehensive fine-grained evaluation of VLT trackers.  For more data analysis of the selected datasets, please refer to Appendix A.4.

The representative datasets in VLTVerse include OTB99\_Lang \cite{otb99}, TNL2K \cite{tnl2k}, LaSOT \cite{lasot}, and MGIT \cite{mgit}. In VLTVerse, OTB99\_Lang and TNL2K serve as representative datasets for short-term tracking. OTB99\_Lang enhances each sequence with semantic information by providing a textual description for the first frame. TNL2K, designed specifically for VLT tasks, includes two thousand video sequences with diverse attributes, setting higher challenges for tracking performance compared to OTB99\_Lang. LaSOT represents long-term tracking, where textual descriptions focus solely on object appearance, excluding positional information. Lastly, MGIT addresses global instance tracking by offering multi-level textual granularity for each sequence, thereby supporting a robust evaluation across various tracking scenarios.

\subsection{Challenge Factor Selection}
In the SOT task, addressing various challenge factors is essential for enhancing tracker performance and robustness. However, current VLT trackers often overlook this aspect. To address this gap, VLTVerse treats the influence of challenge factors in different contexts as a key observation point.

SOTVerse \cite{sotverse} standardizes attribute calculations and defines challenge factors based on their distributions. Following this approach, we categorize attributes into two types: (1) static attributes, which pertain solely to the current frame, and (2) dynamic attributes, which capture changes between consecutive frames. The four static attributes ($c_1$ - $c_4$) are \textit{Abnormal Ratio}, \textit{Abnormal Scale}, \textit{Blur Bounding-box}, and \textit{Abnormal Illumination}. The six dynamic attributes ($c_5$ - $c_{10}$) include \textit{Delta Illumination}, \textit{Delta Scale}, \textit{Delta Blur}, \textit{Delta Ratio}, \textit{Fast Motion}, and \textit{Correlation Coefficient}.

We calculate attribute values for each frame or between frames and then average them across the sequence to derive sequence-level attribute values. Sequences with values falling within specific challenge thresholds are classified as challenge factor sequences, forming the sequence-level challenge factor space detailed in Table \ref{tab:challengingfactor}. For attribute calculation rules and challenge factor thresholds, please refer to SOTVerse \cite{sotverse} and Appendix A.3.

\subsection{Textual Information Setting}
Most existing VLT datasets are labeled with a single granularity, yet we hypothesize that diverse challenge factor spaces may require different types of information (i.e., natural language) to assist tracking. Recently, new annotation approaches have emerged. For example, DTVLT \cite{dtvlt} provides concise and detailed text for the first frame and every hundred frames of each video sequence, while \cite{vlt_ost} explores the impact of attribute words on VLT. However, \textbf{\textit{does text truly enhance tracking when transitioning from SOT to VLT? }}

In VLTVerse, as outlined in Table \ref{tab:information}, we introduce six types of semantic information ($i_1$ - $i_6$): \textit{Attribute Words}, \textit{Initial Concise}, \textit{Dense Concise}, \textit{Initial Detailed}, \textit{Dense Detailed}, and \textit{Blank}. The combinations of textual information $E_i$ and challenge factors $E_c$ create 60 configurations $E_{ci}$, enabling a fine-grained evaluation to reveal the role of language in VLT.

\cite{vlt_ost} provides attribute words for OTB99\_Lang, LaSOT and TNL2K. Based on this, we provide attribute words for MGIT to construct \textit{Attribute Words} space. Similar to DTVLT \cite{dtvlt}, VLTVerse includes \textit{Initial Concise}, \textit{Dense Concise}, \textit{Initial Detailed}, and \textit{Dense Detailed} descriptions. For the \textit{Blank} condition, “The tracking target” is used as a control during tracking.

\begin{itemize}
    \item \textbf{Attribute Words.} Provides four descriptive terms for each object in the video, covering major category, root category, color, and initial position.
    \item \textbf{Initial Concise and Initial Detailed.} The initial concise description includes fundamental details, such as category and initial position, while the initial detailed description expands on this with additional spatio-temporal information, including relative position and initial actions. 
    \item \textbf{Dense Concise and Dense Detailed.} These descriptions update every 100 frames, with dense, concise offering essential information and dense details providing comprehensive spatio-temporal context. This update frequency is designed to sustain the algorithm's memory state and enhance tracking performance \cite{dtvlt,human,humanmemory,memory}.
    \item \textbf{Blank.} Acts as a control, allowing for observation of tracking performance in the absence of text information.
\end{itemize} 

\begin{figure*}[htbp!]
  \centering
    \includegraphics[width=1\linewidth]{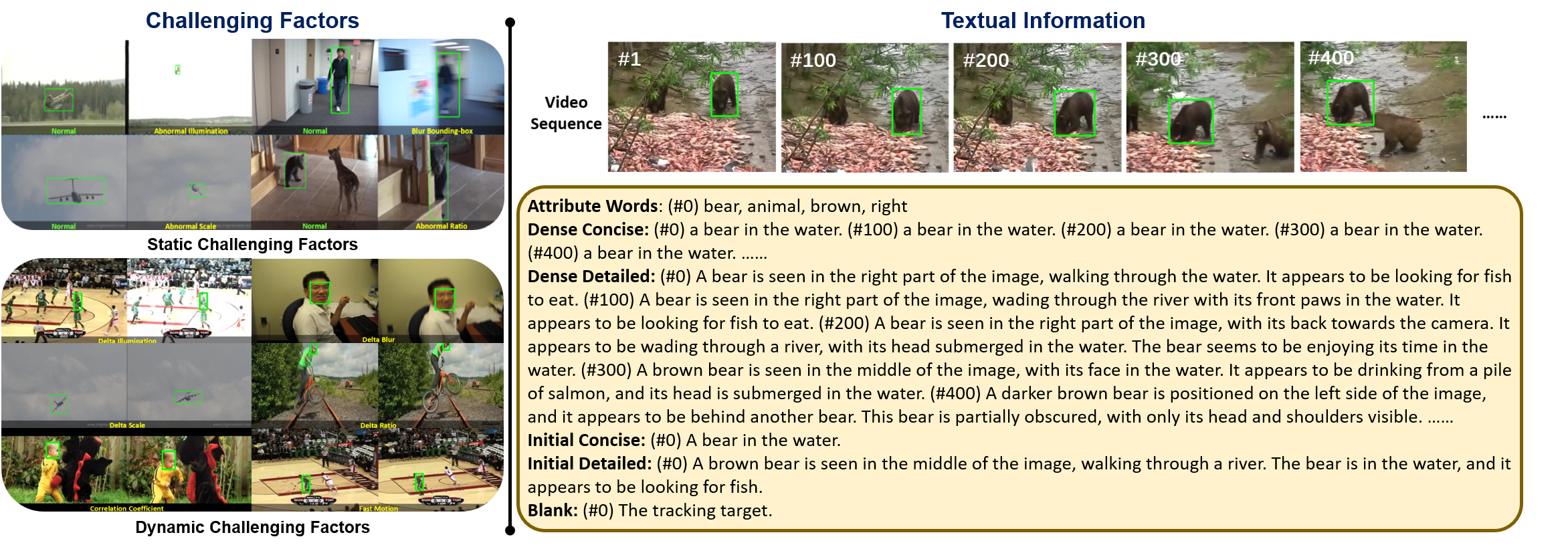}
    \caption{\textbf{Left}: Example of challenging factors, with four static challenging factors and six dynamic challenging factors. \textbf{Right}: Example of textual information, providing six types of information for each video sequence, including Attribute Words, Dense Concise, Dense Detailed, Initial Concise, Initial Detailed, and Blank information.}
    \label{fig:challenge_information_demo}
\end{figure*}

VLTVerse establishes a novel environment for the fine-grained evaluation of VLT, incorporating 10 challenging factors, 6 types of semantic information, and their 60 possible combinations. This framework provides a robust foundation for analyzing the role of language in VLT and identifying potential limitations across various challenging conditions. The visualization of these challenge factors and semantic information types is shown in Figure \ref{fig:challenge_information_demo}.
\section{Evaluation}
\label{sec:evaluation}

\subsection{Evaluation System}
We use the OPE system to evaluate the tracker’s performance on VLTVerse. The evaluation metrics include Area Under the Curve (AUC), Precision score (PRE), Success Score (SUC), Success Rate (SR), and Normalized Precision score (N-PRE). More details about metrics in VLTVerse, please refer to Appendix B.1.

\subsection{Evaluation Settings}
Using VLTVerse, we first evaluate tracker performance across 50 experimental groups, combining 10 challenging factors with 5 types of textual information (excluding Blank as control). This step assesses tracker responses to different challenging factors with varied textual guidance. Subsequently, we analyze the impact of introducing five types of semantic information by comparing results with the Blank control across different trackers and challenge conditions.

To assess performance across datasets with varying challenging factor distributions, we apply weighted scores based on the number of challenge factor sequences within each dataset, providing an overall measure of tracker performance across challenge factors and text prompts. For detailed tracker performance on each dataset, refer to Appendix B.2.

Please note that VLTVerse aims to analyze the influence of semantic information under various challenging conditions, which is unsupported by the official VLT dataset annotations. We discuss the performance of three trackers using the official annotations in Appendix B.4.
\section{Executor}
\label{sec:executor}

\subsection{VLT Trackers}
We select MMTrack \cite{mmtrack}, JointNLT \cite{jointnlt}, and UVLTrack \cite{uvltrack} as baseline models for evaluation on VLTVerse. MMTrack is a typical algorithm for the VLT task, redefined as a token generation task. From a unified modeling perspective, it learns vision-language features, which enhances the model's robustness ability. JointNTL is the first to unify tracking and grounding into a single task, adapting to different references in the grounding and tracking processes and improving adaptability to changes in object appearance. UVLTrack supports SOT, VLT, and visual grounding tasks simultaneously with a single parameter set, using a multi-modal contrastive loss to align features into a unified semantic space. These three models represent distinct paradigms for VLT tasks. VLTVerse enables us to evaluate and identify the limitations of these trackers, providing insights to guide future performance improvements.

\subsection{Implementation Details}
To ensure fair evaluation, all experiments were conducted using the original repository’s hyper-parameters on identical RTX-3090 GPUs. Dense textual information was updated dynamically every hundred frames, while attribute words and initial information were provided only for the first frame. For UVLTrack \cite{uvltrack}, we used the UVLTrack-B model.

\begin{figure*}[htbp!]
  \centering
  \vspace{-20pt}
    \includegraphics[width=1\linewidth]{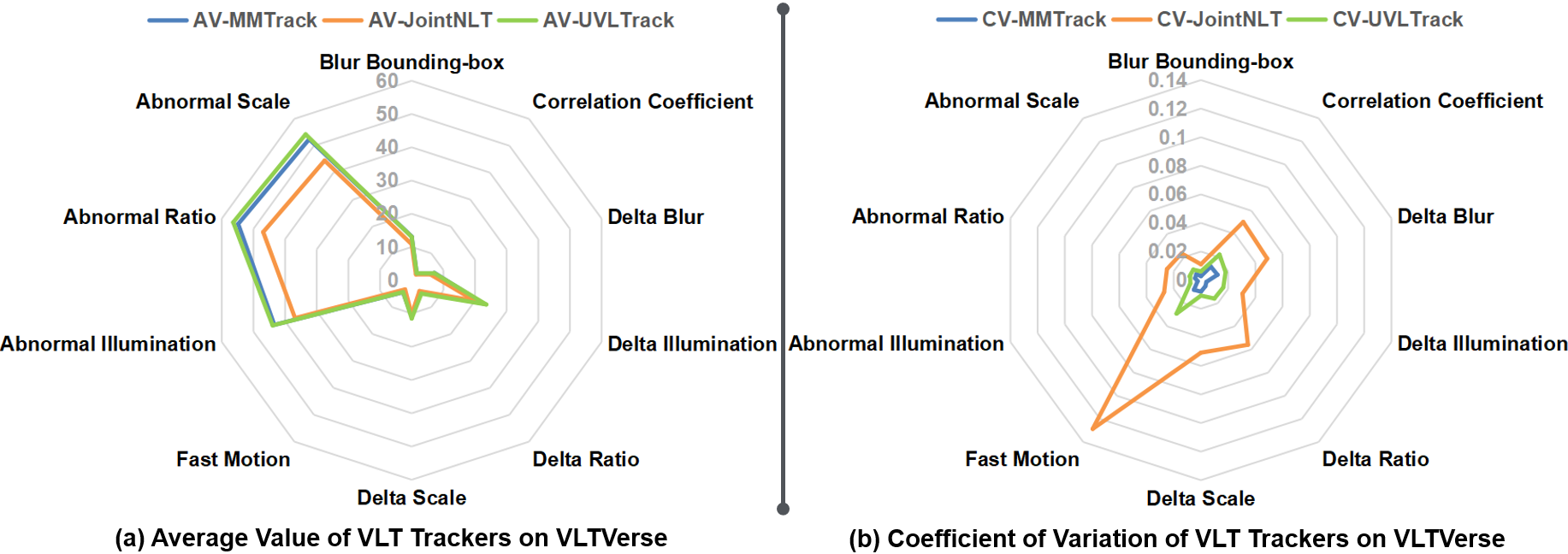}
    \caption{Radar chart of the Average Value (AV) and Coefficient of Variation (CV) of tracking performance under different textual information guidance for various challenging factors (based on SUC).}
    \label{fig:main_results}
\end{figure*}

\section{Experimental Results}
\label{sec:analysis}

\subsection{Challenge Factors \& Meaningful Information}
\begin{figure}[htbp!]
  \centering
  \vspace{-10pt}
   \includegraphics[width=1\linewidth]{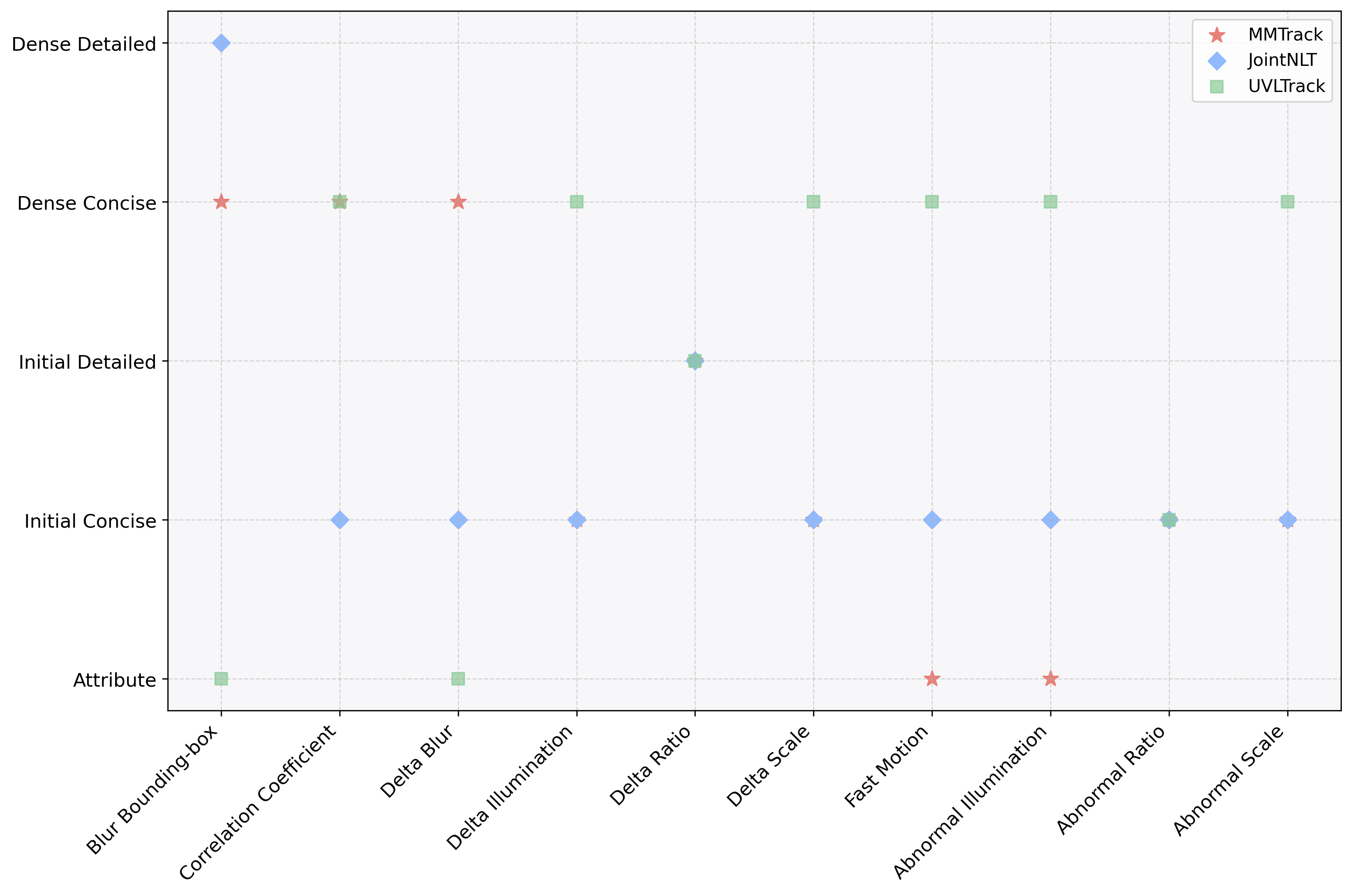}
   \vspace{-8pt}
   \caption{Optimal semantic information for tracking performance under various challenge factors. We use a star, diamond, and square to indicate the best textual information for MMTrack \cite{mmtrack}, JointNLT \cite{jointnlt}, and UVLTrack \cite{uvltrack}, respectively.}
   \label{fig:best_results}
   \vspace{-12pt}
\end{figure}

Figure \ref{fig:main_results} presents radar charts of the Average Value (AV) and Coefficient of Variation (CV) for the performance of three algorithms across various challenge factors and types of textual information (based on SUC). Results for other metrics are available in Appendix B.3. 
The best results for each combination of challenge factors and semantic information are highlighted in Figure \ref{fig:best_results}. From these figures, we can draw the following conclusions:

\textbf{Influence of Challenging Factors.} As shown in Figure \ref{fig:main_results} (a), the most challenging factors for VLT trackers are, in order, correlation coefficient, delta ratio, and fast motion, with the lowest average performance observed under these conditions. Conversely, all trackers perform best under abnormal ratio, abnormal scale, and abnormal illumination. This indicates that challenging factors related to dynamic attributes, such as fast motion and delta ratio, are particularly difficult for trackers to handle. This aligns with expectations, as dynamic challenges present significant obstacles to improving tracker performance from SOT to VLT, whereas static attributes generally pose less difficulty.

\textbf{Influence of Textual Information Based on Challenging Factors.} Due to differences in model design and other factors, trackers show varying adaptability to textual information under challenging conditions. As shown in Figure \ref{fig:main_results} (b), more challenging factors, such as fast motion and delta ratio, are associated with high CV values, indicating greater variability in tracker performance based on textual input. In contrast, simpler challenging factors result in relatively low CV values, suggesting more stable tracking performance across different textual inputs. This indicates that in complex scenarios, different types of textual information significantly affect tracker performance, leading to fluctuations, whereas in simpler scenarios, textual information consistently supports tracking. These findings support the purpose of VLTVerse: to enhance tracker performance by systematically evaluating the role of language in VLT. To achieve this, it is essential to assess tracking performance across various challenging factors and analyze which types of text are most beneficial under these conditions. Such targeted insights can guide the development of strategies for effectively integrating textual information to improve tracking accuracy.

\begin{figure*}[htbp!]
  \centering
  \vspace{-10pt}
   \includegraphics[width=1\linewidth]{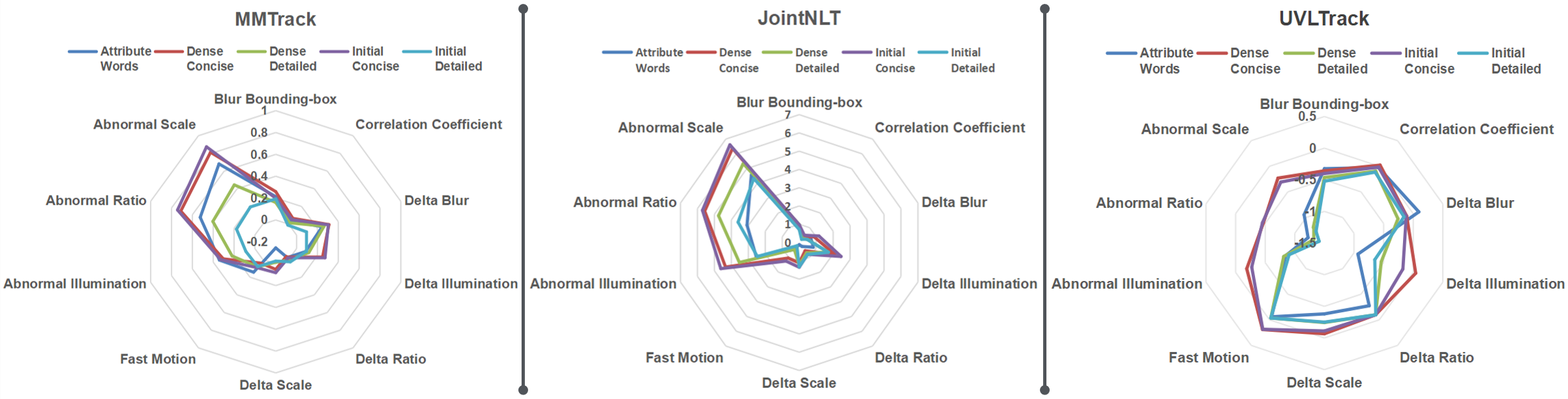}
   \vspace{-8pt}
   \caption{Radar chart of the tracking performance difference in information across various texts versus Blank under different challenge factors (based on SUC).}
   \label{fig:blank_control}
   \vspace{-10pt}
\end{figure*}

\textbf{Influence for different VLT trackers.} Figure \ref{fig:best_results} illustrates the optimal text types for each tracker under various challenging factors. JointNLT \cite{jointnlt} performs best with initial concise information under the most challenging conditions, while UVLTrack \cite{uvltrack} performs optimally with dense, concise information. MMTrack \cite{mmtrack}, however, lacks a consistent pattern across different challenging factors.

We attribute this to JointNLT’s truncation of long texts, which limits its capacity to process lengthy descriptions, resulting in better performance with shorter texts. Additionally, we found that it is very sensitive to different types of text, showing the greatest performance fluctuations when text length and semantics change (highest CV). As a result, during text updates, it fails to benefit from the updates, and some new text also interferes with tracking. This exposes the shortcomings in the design of the current tracker. The tracker should have a good understanding of different text information. The experimental results show that JointNLT uses an approach similar to “memorizing answers” to complete VLT tasks, which contradicts the original intent of the task. This phenomenon does not improve even under different challenging factors. 

UVLTrack uses both SOT and VLT data during training, which enhances the model’s robustness while unifying the two tasks. In situations where tracking through the visual modality is relatively stable, simple text is sufficient to provide necessary information. Compared to JointNLT, UVLTrack is more adaptable to text updates and can benefit from them, making Dense Concise information the most helpful. It is particularly noteworthy that under the static challenge factor Blue Bounding-box and the dynamic challenge factor Delta Blue, UVLTrack only requires attribute information to achieve good tracking performance. This reflects the model’s robustness, indicating that when visual modality tracking is stable enough, it doesn’t need too much textual information for support. 

\begin{figure}[htbp!]
  \centering
  \vspace{-5pt}
   \includegraphics[width=1\linewidth]{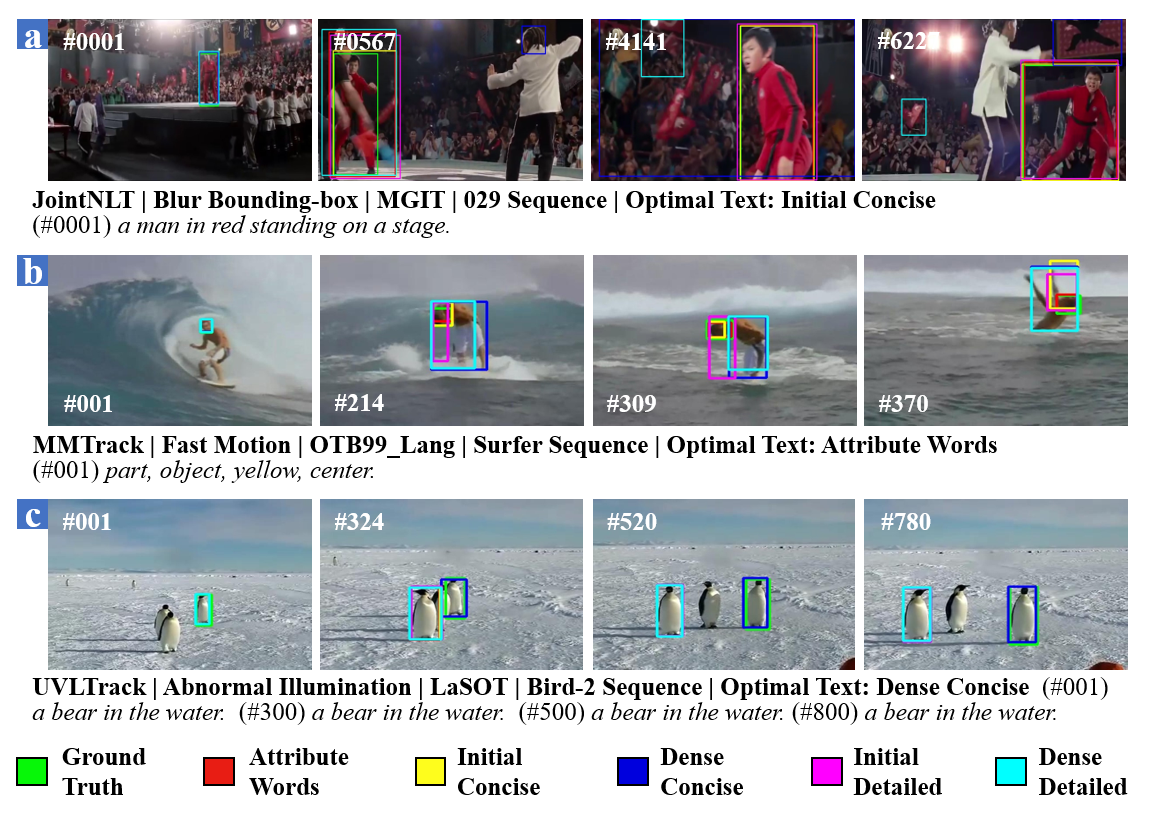}
   \caption{Visualization of tracker results combined with textual information under different challenge factors. We visualize the tracking results of three trackers under representative challenge factors, and due to space constraints, only the best-performing textual information is presented. We use VLTVerse, a fine-grained evaluation framework to reveal the role of language in VLT.}
   \label{fig:visualization}
   \vspace{-10pt}
\end{figure}

MMTrack transforms the VLT task into a token generation task through unified token learning, which helps enhance the learning of visual and language modalities. Unlike JointNTL and UVLTrack, MMTrack shows significant differences under various challenge factors. For Fast Motion and Abnormal Illumination, attribute words provide the greatest assistance. The attribute words describe the inherent properties of the target, which are usually unrelated to lighting conditions and the target's motion state, offering robust clues for the tracker to complete tracking. Unlike UVLTrack, MMTrack requires dense text to provide more guidance for tracking under the Blur Bounding-box and Delta Blur challenge factors. We further visualize the performance of the three trackers under specific challenging factor sequences in Figure \ref{fig:visualization}.

\subsection{Blank Information}
\textbf{Influence of Text Introduction Based on Blank Information.} To assess whether text introduction from SOT to VLT tasks genuinely improves tracking, we use the meaningless text “The tracking target” as input. Figure \ref{fig:blank_control} illustrates the performance differences (based on SUC) between various text types and Blank information across different challenging factors.

As shown in Figure \ref{fig:blank_control}, for MMTrack \cite{mmtrack} and JointNLT \cite{jointnlt}, the performance differences are generally positive, indicating that text introduction enhances tracking performance and robustness under challenging conditions. 
However, for UVLTrack \cite{uvltrack}, performance differences are mostly negative, suggesting that meaningless text may interfere with tracking. This is likely due to UVLTrack’s architecture, which uses unified parameters for SOT, VLT, and visual grounding tasks. When provided with irrelevant text, UVLTrack can rely on visual cues alone to achieve stable tracking, revealing that it has not fully leveraged the potential benefits of text.
\section{Conclusion}
\label{sec:conclusion}

In this paper, we introduce VLTVerse, a fine-grained evaluation framework that illuminates the role of language in VLT tasks. By expanding traditional evaluation methods, VLTVerse provides 60 unique combinations of 10 challenge factors and 6 semantic information types to assess VLT trackers. Using this framework, we perform an in-depth evaluation of three mainstream VLT trackers—MMTrack, JointNLT, and UVLTrack—identifying key performance bottlenecks associated with specific challenge factors and text types. Our analysis reveals which challenging factors most significantly impact tracker performance and robustness, as well as how varying text inputs can lead to performance fluctuations. For example, shorter texts benefit JointNLT due to its limited capacity for long text, while dense text supports MMTrack under blur conditions. Additionally, we find that certain trackers, such as UVLTrack, often perform well even with minimal text, highlighting room for optimizing text integration. We hope that VLTVerse provides actionable insights for improving VLT trackers from the perspectives of data, evaluation, and algorithm design, thereby advancing tracking robustness and accuracy.

{
    \small
    \bibliographystyle{ieeenat_fullname}
    \bibliography{main}
}


\end{document}